\documentclass[10pt,twocolumn,letterpaper]{article}

\usepackage{cvpr}              
\definecolor{cvprblue}{rgb}{0.21,0.49,0.74}
\usepackage[pagebackref,breaklinks,colorlinks,allcolors=cvprblue]{hyperref}

\newcommand{\Eq}{\mathbb{E}_{(V,q)}}
\newcommand{\Ei}{\mathbb{E}_{i}}
\newcommand{\KL}{D_{\mathrm{KL}}}
\newcommand{\clipeps}[1]{\operatorname{clip}_{\epsilon}\!\left(#1\right)} 

\usepackage{amsmath}
\usepackage{booktabs}
\usepackage{adjustbox}
\usepackage[table]{xcolor}
\usepackage{makecell}
\usepackage{arydshln}
\usepackage{colortbl}

\usepackage[accsupp]{axessibility} 

\def\paperID{101} 
\def\confName{CVPR}
\def\confYear{2026}

\title{Reinforce to Learn, Elect to Reason: A Dual Paradigm for Video Reasoning}

\author{Songyuan Yang$^{1*}$ \quad Weijiang Yu$^{2\dagger}$ \quad Jilin Ma$^2$ \quad Ziyu Liu$^2$ \quad Guijian Tang$^1$ \\ Wenjing Yang$^{1}$ \quad Huibin Tan$^{1*}$ \quad Nong Xiao$^2$ \\
$^1$National University of Defense Technology \qquad $^2$ Sun Yat-sen University
}

\begin{document}
\maketitle
\begin{abstract}
Video reasoning has advanced with large multimodal models (LMMs), yet their inference is often a single pass that returns an answer without verifying whether the reasoning is evidence-aligned. We introduce Reinforce to Learn, Elect to Reason (RLER), a dual paradigm that decouples learning to produce evidence from obtaining a reliable answer. In RLER-Training, we optimize the policy with group-relative reinforcement learning (RL) and 3 novel task-driven rewards: Frame-sensitive reward grounds reasoning on explicit key frames, Think-transparency reward shapes readable and parsable reasoning traces, and Anti-repetition reward boosts information density. These signals teach the model to emit structured, machine-checkable evidence and potentiate reasoning capabilities. In RLER-Inference, we apply a train-free orchestrator that generates a small set of diverse candidates, parses their answers and cited frames, scores them by evidence consistency, confidence, transparency, and non-redundancy, and then performs a robust evidence-weighted election. This closes the loop between producing and using evidence, improving reliability and interpretability without enlarging the model. We comprehensively evaluate RLER against various open-source and RL-based LMMs on 8 representative benchmarks. RLER achieves state of the art across all benchmarks and delivers an average improvement of 6.3\% over base models, while using on average 3.1 candidates per question, indicating a favorable balance between compute and quality. The results support a simple thesis: making evidence explicit during learning and electing by evidence during inference is a robust path to trustworthy video reasoning.

\makeatletter{\renewcommand*{\@makefnmark}{}
\footnotetext{$^*$Equal contribution.
\textsuperscript{$^\dagger$}Corresponding author.}
}

\end{abstract}    
\begin{figure}[t]
    \centering
    \includegraphics[width=\linewidth]{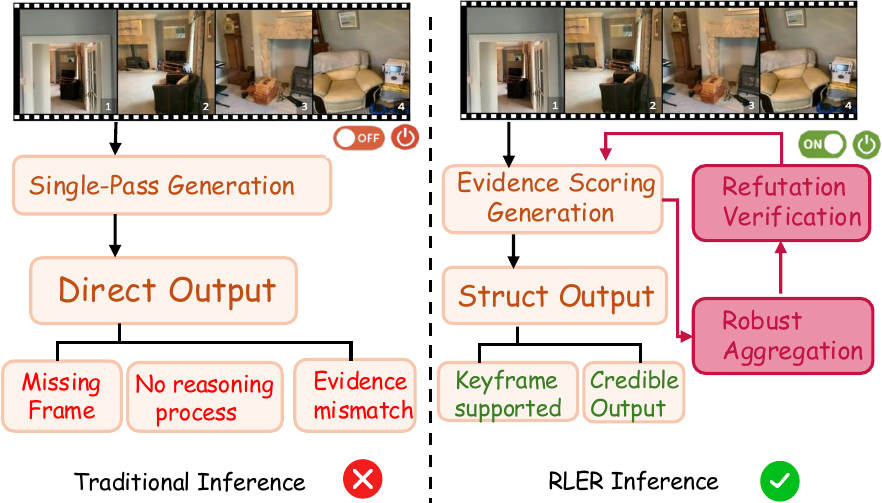}
    \vspace{-6mm}
    \caption{\textbf{Comparison between Traditional and RLER Inference.} Single pass inference outputs an answer without verification and may miss key evidence. RLER produces structured outputs, scores evidence across candidates, aggregates by evidence, and performs a refutation check to deliver credible answers.}
    \vspace{-6mm}
    \label{fig:introduction}
\end{figure}

\vspace{-4mm}
\section{Introduction}
\label{sec:intro}

The rapid evolution of large multimodal models (LMMs) is pushing video understanding from seeing to reasoning \cite{zhang2023videollama, liu2023llava, liu2024llavanext} and has enabled stronger cross-frame and open-domain inference \cite{openai2024openaio1card, shao2024visual, feng2025videor1}. However, a persistent weakness remains in practice \cite{chen2025cgbench, liu2025commonsense}: inference is often single pass, returning an answer without checking whether the reasoning is grounded in evidence or follows an appropriate path. Even sota LMMs can be disrupted by tiny perturbations at inference \cite{xiao2025videoqa, jung2025consistency}, such as minor video changes or alternative phrasings. These vulnerabilities are further amplified by the long temporal horizons and multi-view noise of video \cite{zou2024secondshoursreviewingmultimodal, ICLR2025_f7b77476}. Therefore, during such a single-pass reasoning process, this fragility can accumulate without being detected or corrected, leading to conclusions that are unreliable and difficult to interpret in complex video reasoning.

Recent work improves reasoning through reinforcement-style training with rule-based rewards, format constraints, and spatio-temporal guidance \cite{deepseekr1, zhan2025visionr1, li2025star}, and through test-time scaling with diversified sampling or search \cite{gui2024bonbon, xie2023self, feng2023alphazero}. These researches bring measurable gains, but a critical gap remains: multiple reasoning traces are rarely checked for evidence consistency, chain-of-thought is rarely verified against key frames and relations in a systematic way, and there is no principled arbitration across traces. In short, as shown in Figure~\ref{fig:introduction}, prior work often shows that a model can reason, but not that it reasoned using the correct evidence, which defines the challenge and our point of departure.

Our motivation is to move video reasoning from answer-driven to evidence-driven. Specifically, training serves as shaping and potentiation: it teaches the model to emit structured, machine-readable evidence signals. In contrast, reliability is decided at inference by whether the system can elect a consistent outcome from that evidence, aggregate it robustly, and perform a minimal self-review when needed. We therefore propose the dual paradigm RLER (Reinforce to Learn, Elect to Reason). Reinforce Learning shapes explicit evidence and potentiates reasoning capabilities, while inference makes decisions by evidence rather than by a single chain, which separates ``being able to think" from ``thinking correctly" and closes the loop in one framework.

On the training side (RLER-Training), we adopt GRPO \cite{grpo} and design three novel rewards that target the essence of video reasoning: where to look, how much to say, and how to say it. The Frame-sensitive Reward encourages the identification of explicit key frames and their citation within the reasoning text, promoting cross-frame relations and temporal dependencies. The Anti-repetition Reward suppresses low-information repetition to increase information density and provide cleaner signals for subsequent comparison. The Think-transparency Reward encourages more elaborate and interpretable thought processes. Combined with the conventional Formatted Reward and Accuracy Reward, these novel rewards yield evidence-centric, readable, and verifiable trajectories that interface directly with inference-time orchestration.

On the inference side (RLER-Inference), we introduce a train-free evidence-aligned orchestrator. For each video–question pair,  the system first performs diversified inputs and decoding across styles while keeping one conservative run. This produces a set of independent candidate thoughts and answers. We parse each candidate into answer, key frame citations, and reasoning, then compute an evidence score from evidence co occurrence and relational consistency and combine it with length normalized confidence, redundancy, and transparency. Aggregation removes extreme candidates and performs a weighted election by evidence; if the top margin is insufficient the sample budget expands, and if an evidence weighted majority with sufficient confidence is met the process stops early. A single referee style self check verifies that cited frames and logic support the winner; if stronger counter evidence is found we reweight and draw one extra sample, otherwise we confirm the answer. This turns multi sample generation into an evidence centered election and review, making the shaped evidence a computable, early stopping, and correctable decision mechanism. Extensive experiments demonstrate the effectiveness of our approach.

Our contributions are as follows: 
(1) We present the dual paradigm RLER (Reinforce to Learn, Elect to Reason) that closes the loop between reward-shaped training and evidence-aligned election at inference, addressing the unreliability of single-pass generation in video reasoning.
(2) We design 3 novel training rewards, Frame-sensitive Reward, Think-transparency Reward, and Anti-repetition Reward, which ensure evidence-centric and structured reasoning outputs and align naturally with inference-time orchestration.
(3) We propose a train-free inference orchestrator (RLER-Inference) that integrates diversified sampling, evidence extraction and scoring, evidence-weighted election with outlier removal, adaptive budgeting with early stopping, and a one-shot referee-style self-check, leading multiple thoughts to a stable consensus at the level of evidence.
(4) Extensive experiments demonstrate that RLER is broadly applicable and cost-effective, yielding an average improvement of 6.3\% across benchmarks while using about 3.1 candidates per question, indicating strong effectiveness.

\begin{figure*}[t]
    \centering
    \includegraphics[width=\linewidth]{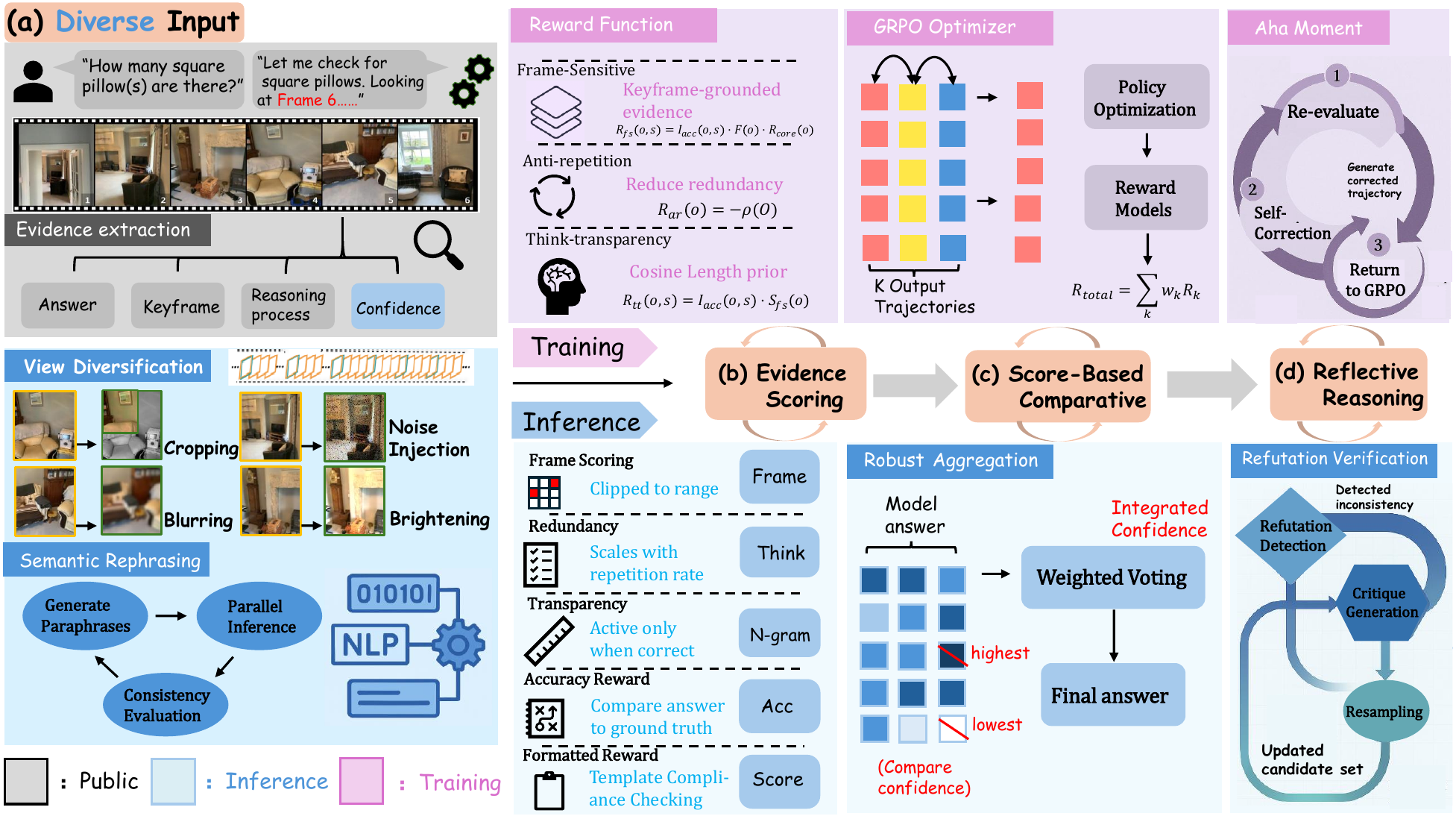}
    \vspace{-6mm}
    \caption{\textbf{Method Overview.} RLER-Training uses GRPO with Frame-sensitive, Think-transparency, and Anti-repetition Rewards to teach the model to emit structured outputs with keyframe citations and potentiate reasoning capabilities. RLER-Inference applies diverse inputs to produce multiple candidates, parse structure, score evidence, aggregate by evidence weights, and run refutation check.}
    \label{fig:method}
    \vspace{-5mm}
\end{figure*}
\vspace{-1mm}
\section{Related Works}

\noindent\textbf{Large Multimodal Models.}
Large Multimodal Models (LMMs) extend language models \cite{yi, llama4, qwen3, gpt4} to visual \cite{liu2023llava, liu2024llavanext, lan2023learning, cheng2024videollama2} modality, enabling unified perception–language understanding across images and videos. Early work established strong results on captioning and visual question answering, and recent research focus has been shifted towards extending LMMs to handle video understanding, with methods such as Qwen-2.5VL \cite{Qwen2.5-VL}, InternVL2.5 \cite{chen2024internvl2_5}, LLaVA-Video \cite{zhang2024llavavideo} and Video-LLaMA 3 \cite{zhang2025videollama3}. However, most LMMs still produce answers in a single pass without verifiable evidence alignment, which our RLER framework addresses by learning to emit structured evidence and electing the final answer by that evidence at inference.

\noindent\textbf{Video Reasoning.}
Video reasoning has advanced along two complementary axes: training and inference. On the training side, reinforcement learning with verifiable or rule based rewards, exemplified in text by OpenAI o1 \cite{openai2024openaio1card} and DeepSeek-R1 \cite{deepseekr1}, has been adapted to visual settings through task specific reward designs that strengthen fine grained perception and reasoning \cite{zhou2025r1,peng2025lmm,visualrft,r1vl,zhan2025visionr1,deng2025openvlthinker}. Although video understanding is less explored, recent efforts indicate promise for temporal and causal reasoning in videos \cite{feng2025videor1, yan2025videochatr15, wang2025videorft, li2025star}. On the inference side, test time scaling methods such as best of N sampling \cite{gui2024bonbon}, guided beam search \cite{xie2023self}, and Monte Carlo tree search \cite{feng2023alphazero, li2025dyfo, zhang-etal-2025-vrest} improve reasoning without additional training, yet applications to LMMs largely mirror text only heuristics, including response lengthening \cite{shao2024visual} or search based reranking \cite{wang2025videotree}. However, these two lines are often applied in isolation and seldom enforce cross candidate evidence consistency; our RLER unifies reward shaping with evidence aligned inference to address this gap.

\vspace{-1mm}

\section{Preliminaries}

\subsection{Formalization for Video Reasoning}
Let a video be $V=(x_{1},\ldots,x_{T})$ with frames $x_{t}$ and length $T$. Given a question $q$, a policy model $\pi_{\theta}$ generates a reasoning trace $z$ and an answer $a$. We write $y=(z,a)$ and denote keyframe indices extracted from $z$ as $K \subseteq {1,\ldots,T}$. The supervised corpus provides triplets $(V,q,a^{\star})$, while the policy may produce multiple candidates $y$ per $(V,q)$. A scalar reward $r(y;V,q)$ evaluates $y$ with respect to the task. In our framework, the training stage RLER-Training will instantiate $r$ with task-specific rewards and optimize $\pi_{\theta}$; the inference stage RLER-Inference will later use the structured trace $(z,K)$ for evidence-aligned selection. 

\subsection{Group Relative Policy Optimization}
Group Relative Policy Optimization (GRPO)~\cite{grpo,deepseekr1} is an on-policy method that optimizes $\pi_{\theta}$ without a learned critic. For each $(V,q)$, the algorithm samples a group of $G$ candidates $\{o_i\}_{i=1}^{G}$ from the old policy $\pi_{\theta_{\text{old}}}(\cdot\,|\,V,q)$ and computes rewards $r_{i}=r(o_{i};V,q)$. A normalized advantage is obtained by a z-score within the group
\begin{equation}
A_{i}=\frac{r_{i}-\bar{r}}{\sigma_{r}+\epsilon},\qquad
\bar{r}=\tfrac{1}{G}\sum_{j=1}^{G}r_{j},\quad
\sigma_{r}^{2}=\tfrac{1}{G}\sum_{j=1}^{G}(r_{j}-\bar{r})^{2}.
\label{eq:grpo-adv}
\end{equation}
Let the probability ratio be
\begin{equation}
\rho_{i}(\theta)=\frac{\pi_{\theta}(o_{i}\,|\,V,q)}{\pi_{\theta_{\text{old}}}(o_{i}\,|\,V,q)}.
\label{eq:grpo-ratio}
\end{equation}
GRPO maximizes a clipped surrogate with a KL regularizer, analogous to PPO~\cite{schulman2017ppo}, but using group-normalized advantages:
\begin{equation}
\label{eq:grpo-clip}
\overline{\mathcal L}_{\text{clip}}(\theta)
= \Eq\!\Big[\Ei\big[\min\big(\rho_i A_i,\, \clipeps{\rho_i} A_i\big)\big]\Big].
\end{equation}
\begin{equation}
\label{eq:grpo-reg}
\mathcal{R}_{\mathrm{KL}}(\theta)
= \Eq\!\big[\KL\!\left(\pi_{\theta}\,\|\,\pi_{\text{ref}}\right)\big].
\end{equation}
\begin{equation}
\label{eq:grpo-obj}
\mathcal J_{\text{GRPO}}(\theta)
= \overline{\mathcal L}_{\text{clip}}(\theta) - \beta\,\mathcal{R}_{\mathrm{KL}}(\theta).
\end{equation}
The first term encourages responses with above-average rewards within the sampled group while preventing unstable updates through clipping. The second term penalizes divergence from a fixed reference policy $\pi_{\text{ref}}$ with weight $\beta$, which stabilizes training and limits drift. Parameters are updated by stochastic gradient ascent on $\mathcal{J}_{\text{GRPO}}(\theta)$ with periodic refresh of $\pi_{\theta_{\text{old}}}$. All probabilities in Eq.~\eqref{eq:grpo-ratio}–\eqref{eq:grpo-obj} are conditioned on $(V,q)$, and $\mathbb{E}_{(V,q),\,i}$ averages over data pairs and a uniform index $i\in\{1,\ldots,G\}$. In RLER-Training, this objective is combined with the task reward to shape evidence-centric outputs that are consumed by RLER-Inference at inference stage.

\section{Reinforce to Learn, Elect to Reason}
Reinforce to Learn, Elect to Reason (RLER) is a dual paradigm that first learns to produce evidence and then reasons by electing among it, as shown in Figure~\ref{fig:method}. RLER-Training uses reinforcement to shape structured, verifiable signals and to strengthen the model’s reasoning capacity, making spatiotemporal grounding explicit and encouraging disciplined chains of thought. RLER-Inference consumes these signals, compares multiple candidate thoughts, aligns them with evidence, and performs self-review before deciding. The split is deliberate: learning determines where to look, what to say, and how to express it in a machine-parsable form; inference determines how to aggregate, arbitrate, and verify. Both stages adopt the same evidence-centric view, creating a natural symmetry between training rewards and inference-time scoring. 

\subsection{RLER-Training}
We optimize the policy with GRPO using rewards that shape evidence-centric, concise, and transparent reasoning. For a video–question pair $(V,q)$, let $o$ denote the model output and $s$ the ground-truth answer. The total reward aggregates task components with nonnegative weights that sum to one,
\begin{equation}
R_{\text{total}}(o,s)=\sum_{k} w_k\, R_k(o,s),\qquad w_k\ge 0,\ \sum_k w_k=1.
\label{eq:RLER-total}
\end{equation}
Two indicators are shared across rewards: $\mathbb{I}_{\text{acc}}(o,s)\in\{0,1\}$ signals answer correctness, and $\mathcal{F}(o)\in\{0,1\}$ signals that the output follows the required schema with \texttt{<think>}, \texttt{<answer>}, and \texttt{<keyframes>} tags.

\noindent\textbf{Frame-sensitive Reward.}
This term links the reasoning trace to explicit visual evidence. Let $K(o)$ be the set of valid keyframe indices extracted from $o$, and let $E(o)$ count invalid indices. We define a bounded, parameter-light score that grows with valid citations and decreases with invalid ones,
\begin{equation}
s_{\text{fs}}(o)=\operatorname{clip}\!\left(\frac{|K(o)|-E(o)}{1+|K(o)|},\,0,\,1\right).
\label{eq:kqs-compact}
\end{equation}
The final reward is gated by accuracy and format validity,
\begin{equation}
R_{\text{fs}}(o,s)=\mathbb{I}_{\text{acc}}(o,s)\,\mathcal{F}(o)\,s_{\text{fs}}(o).
\label{eq:fs-final}
\end{equation}
This formulation encourages citing informative frames when the model solves the task and adheres to the schema, while providing diminishing returns for excessive frame lists.

\noindent\textbf{Think-transparency Reward.}
We encourage structured and readable traces without promoting verbosity. Let $L(o)$ be the length of the reasoning segment and let $\widetilde{L}(o)\in[0,1]$ be its robustly normalized value obtained by clipping $L(o)$ to a fixed operating range and rescaling. A simple unimodal bonus prefers moderate lengths,
\begin{equation}
R_{\text{tt}}(o,s)=\mathbb{I}_{\text{acc}}(o,s)\,\sin^{2}\!\Bigl(\pi\,\widetilde{L}(o)\Bigr).
\label{eq:tt-final-compact}
\end{equation}
The peak near mid-range discourages both terse shortcuts and unnecessarily long narratives.

\noindent\textbf{Anti-repetition Reward.}
To increase information density, we penalize repeated surface forms. Using fixed small $n$-grams \cite{brown1992ngram}, let $\mathcal{U}_n(o)$ be the set of unique $n$-grams and let $T_n(o)$ be the total count. The repetition ratio and reward are
\begin{equation}
\rho(o)=
\begin{cases}
1-\dfrac{|\mathcal{U}_n(o)|}{T_n(o)}, & T_n(o)>0,\\
0, & \text{otherwise,}
\end{cases}
\qquad
R_{\text{ar}}(o)=-\,\rho(o).
\label{eq:ar-final-compact}
\end{equation}
This pushes the model to avoid vacuous loops and to concentrate on novel evidence.

\noindent\textbf{Formatted Reward.}
To ensure reliable parsing during inference, we assign
\begin{equation}
R_{\text{fmt}}(o)=\mathcal{F}(o),
\label{eq:fmt-compact}
\end{equation}
so malformed outputs receive no structural credit.

\noindent\textbf{Accuracy Reward.}
Task success remains the principal signal,
\begin{equation}
R_{\text{acc}}(o,s)=\mathbb{I}_{\text{acc}}(o,s).
\label{eq:acc-compact}
\end{equation}
The five components in Eq.~\eqref{eq:fs-final}–\eqref{eq:acc-compact} are combined by Eq.~\eqref{eq:RLER-total} and optimized with GRPO. The resulting policy emits key frames, structured and concise reasoning, and schema-compliant outputs. These properties are consumed by RLER-Inference, which scores evidence, compares candidates, and performs a brief self-review to select a reliable answer.

\subsection{RLER-Inference}

RLER-Inference consumes the evidence-shaped traces produced by training and turns multi-sample generation into a principled, evidence-aligned election. Given a video–question pair $(V,q)$ and the trained policy $\pi_{\theta}$, we draw a small set of diversified candidates by perturbing decoding or the visual crop while preserving semantics. Formally, let $\xi_i$ denote a light perturbation. We sample
\begin{equation}
o_i \sim \pi_{\theta}^{\,\xi_i}(\cdot\,|\,V,q),\qquad i=1,\ldots,K,
\label{eq:RLER-sample}
\end{equation}
and collect $\mathcal{O}=\{o_i\}_{i=1}^{K}$. Each output follows the schema enforced at training and exposes the fields needed for downstream scoring.

From every $o_i$ we parse a canonical answer, a set of cited key frames, and the reasoning text. We write
\begin{equation}
(a_i,\;K_i,\;z_i,\;c_i)=\phi(o_i),
\label{eq:parse}
\end{equation}
where $a_i$ is the normalized answer string, $K_i\subseteq\{1,\ldots,T\}$ are key-frame indices, $z_i$ is the reasoning segment, and $c_i\in[0,1]$ is a confidence proxy derived from token probabilities over a short tail segment of $o_i$ with length normalization. The map $\phi$ is deterministic and relies only on the schema learned during training.

Evidence-aligned scoring mirrors the training rewards in a parameter-light way. We reuse the frame-sensitive term by setting $s_{\mathrm{fs}}(o_i)$ as in Eq.~\eqref{eq:kqs-compact}. We measure transparent thinking by a bounded length preference,
\begin{equation}
\tau(o_i)=\sin^{2}\!\Bigl(\pi\,\widetilde{L}(o_i)\Bigr),\qquad \widetilde{L}(o_i)\in[0,1],
\label{eq:tau}
\end{equation}
where $\widetilde{L}(o_i)$ is a clipped-and-rescaled version of the reasoning length. We quantify non-redundancy via $1-\rho(o_i)$, with $\rho(o_i)$ defined in Eq.~\eqref{eq:ar-final-compact}. The composite evidence score averages these signals with the confidence proxy,
\begin{equation}
S_i=\tfrac{1}{4}\Big(s_{\mathrm{fs}}(o_i)+\tau(o_i)+\big(1-\rho(o_i)\big)+c_i\Big)\;\in[0,1].
\label{eq:score}
\end{equation}
This score favors candidates that ground on plausible frames, provide readable structure, avoid repetition, and are internally confident, thus aligning inference with the biases learned by RLER-Training.

Aggregation proceeds by evidence-weighted election over canonical answers. Let $\mathrm{canon}(\cdot)$ map free-form answers to canonical strings and let $\mathcal{A}$ be the set of unique canonicals in $\mathcal{O}$. For $\hat{a}\in\mathcal{A}$, define its supporting index set $I(\hat{a})=\{i:\mathrm{canon}(a_i)=\hat{a}\}$. We form a soft consensus of supporting frames,
\begin{equation}
\bar{K}(\hat{a})=\operatorname*{arg\,max}_{K\subseteq\{1,\ldots,T\}}
\sum_{i\in I(\hat{a})}S_i\,\frac{|K\cap K_i|}{|K\cup K_i|},
\label{eq:consensus-K}
\end{equation}
and an evidence-weighted answer score,
\begin{equation}
S(\hat{a})=\sum_{i\in I(\hat{a})}S_i\;\frac{|K_i\cap \bar{K}(\hat{a})|}{|K_i\cup \bar{K}(\hat{a})|}.
\label{eq:answer-score}
\end{equation}
The elected answer is $\hat{a}^{\star}=\operatorname*{arg\,max}_{\hat{a}\in\mathcal{A}}S(\hat{a})$. To avoid outlier influence, we discard at most one maximal and one minimal $S_i$ within each $I(\hat{a})$ before computing Eq.~\eqref{eq:answer-score}. Early stopping is triggered when the margin $S(\hat{a}^{\star})-\max_{\hat{a}\ne \hat{a}^{\star}}S(\hat{a})$ exceeds a small fixed threshold and the mean confidence $\tfrac{1}{|I(\hat{a}^{\star})|}\sum_{i\in I(\hat{a}^{\star})}c_i$ is above a preset level; otherwise the budget expands with a few additional samples in Eq.~\eqref{eq:RLER-sample}.

Before finalizing, RLER-Inference performs a single referee-style self-check that mirrors the self-correction behaviors observed during training. Conditioned on $(V,q)$ and the tuple $(\hat{a}^{\star},\bar{K}(\hat{a}^{\star}))$, the model produces a short critique that only inspects support sufficiency and potential omissions in the cited frames. The critique induces a nonpositive adjustment $\Delta(\hat{a}^{\star})\le 0$ derived from detected support gaps. The final decision is
\begin{equation}
\hat{a}^{\mathrm{final}}=\operatorname*{arg\,max}_{\hat{a}\in\{\hat{a}^{\star},\,\hat{a}_{(2)}\}}
\Big(S(\hat{a})+\mathbf{1}\{\hat{a}=\hat{a}^{\star}\}\,\Delta(\hat{a}^{\star})\Big),
\label{eq:referee}
\end{equation}
where $\hat{a}_{(2)}$ is the runner-up by $S(\cdot)$. If $\hat{a}_{(2)}$ wins in Eq.~\eqref{eq:referee}, a single additional sample is drawn and the aggregation is recomputed; otherwise the system outputs $\hat{a}^{\star}$. This final step closes the duality with RLER-Training: the same notions of frame grounding, transparency, and non-redundancy that shape the traces during learning become quantitative levers for election and light self-review at inference, yielding a reliable answer with minimal extra computation.

\begin{table*}[h]
\centering
\caption{\textbf{Comparison across 8 video benchmarks grouped by task category.} “w/o sub” denotes evaluation without subtitles and “–” denotes not reported. RLER achieves consistent gains over its baseline and representative open-source and RL-based LMMs on most benchmarks. \textbf{Bold} means sota results among open-source LMMs. Scores are \%.}
\vspace{-2mm}
\begin{adjustbox}{width=\linewidth,center}
\renewcommand{\arraystretch}{1.2}
\setlength{\tabcolsep}{1.5mm}
\begin{tabular}{lllllllll}
\toprule
& \multicolumn{2}{c}{\textbf{Video Reasoning}} & \multicolumn{4}{c}{\textbf{General Video Understanding}} & \multicolumn{2}{c}{\textbf{Long Video Understanding}} \\
\cmidrule(lr){2-3}\cmidrule(lr){4-7}\cmidrule(l){8-9}
\textbf{Model} & \textbf{VSIBench} & \textbf{VideoMMMU} & \makecell{\textbf{VideoMME} \\ \textbf{(w/o sub)}} & \textbf{TempCompass} & \textbf{MVBench} & \textbf{WildVideo} & \textbf{LVBench} & \textbf{LongVideoBench} \\
\midrule
\multicolumn{9}{c}{\textbf{\textit{Proprietary LMMs}}} \\
\midrule
GPT-4o \cite{gpt4o} & 34 & 61.2 & 67.9 & - & 57.5 & 62.1 & 30.8 & 66.7 \\
\midrule
\multicolumn{9}{c}{\textbf{\textit{Open-Source LMMs}}} \\
\midrule
LLaVA-OneVision-7B \cite{li2024llavaonevision} & 32.4 & 33.8 & 58.2 & - & 56.7 & - & - & - \\
ShareGPT4Video-8B \cite{chen2024sharegpt4video}  & -    & -    & -    & - & -    & -    & -   & 39.7 \\
LLaVA-Video-7B \cite{zhang2024llavavideo}     & 35.7 & 36.1 & 63.7 & 65.5 & 62.1 & 53.4 & -   & 59.5 \\
VILA-1.5-8B \cite{lin2023vila}        & 28.9 & 20.9 & -    & 58.8 & -    & -    & -   & - \\
VideoLLaMA 3-7B \cite{zhang2025videollama3}    & -    & 46.0 & 61.0 & -    & -    & -    & 45.3 & - \\
MiniCPM-V 2.6-8B \cite{yao2024minicpm}   & -    & -    & 59.7 & 59.6 & 44.7 & 46.4 & 43.5 & - \\
mPLUG-Owl3-8B \cite{ye2024mplug}      & -    & -    & 53.5 & -    & 54.5 & -    & -   & 59.8 \\
InternVL2.5-8B \cite{chen2024internvl2_5}     & 41.6 & -    & 63.7 & 68.7 & 70.5 & -    & -   & - \\
Qwen2.5-VL-7B \cite{Qwen2.5-VL}     & 37.4 & 47.4 & 65.1 & 69.2  & 67.5 & 51.3 & 42.0 & 56.0 \\
\midrule
\multicolumn{9}{c}{\textbf{\textit{RL-based LMMs}}} \\
\midrule
Video-R1 \cite{feng2025videor1}          & 35.8 & 52.3 & 59.3 & 73.2 & 63.9 & - & - & - \\
STAR-R1 \cite{li2025star}           & 34.1 & 49.2 & 56.6 & 72.4 & 67.8 & - & - & - \\
TinyLLaVA-Video-R1 \cite{zhang2025tinyllava} & -    & -    & 46.6 & 49.5 & -    & - & - & - \\
VideoChat-R1 \cite{li2025videochat}       & -    & 50.0 & 58.8 & 73.9 & 67.9 & - & - & - \\
VideoChat-R1.5 \cite{yan2025videochatr15}     & -    & 51.4 & 67.1 & -    & 70.6 & - & 48.4 & - \\
VideoRFT \cite{wang2025videorft}           & 36.8 & 51.1 & 59.8 & 73.7 & 62.1 & - & - & - \\
MOSS-ChatV \cite{tao2025moss}         & -    & 50.2 & 60.0 & 72.9 & 67.6 & - & - & - \\
\midrule
\rowcolor{blue!15} RLER (Ours) & \textbf{43.3} & \textbf{54.2} & \textbf{68.5} & \textbf{76.2} & \textbf{72.9} & \textbf{57.5} & \textbf{50.7} & \textbf{63.0} \\
\bottomrule
\end{tabular}
\end{adjustbox}
\label{tab:video_results_grouped}
\vspace{-5.5mm}
\end{table*}

\section{Experiment}
\subsection{Experimental Setup}
\noindent\textbf{Benchmarks and Metrics.}
We evaluate on eight public video benchmarks: VSIBench \cite{yang2025vsibench} and VideoMMMU \cite{hu2025videommmu} for multi-step reasoning; VideoMME \cite{fu2025video}, TempCompass \cite{liu2024tempcompass}, MVBench \cite{li2024mvbench}, and WildVideo \cite{yang2025wildvideo} for general understanding; LVBench \cite{wang2025lvbench} and LongVideoBench \cite{wu2024longvideobench} for long-horizon understanding. The primary metric is accuracy (\%) under each benchmark’s official protocol. We further report inference-time compute with the average number of evaluated candidates, $\text{Avg }K$, measured after early stopping under a common cap $K_{\max}$. To analyze evidence quality, we introduce three auxiliary metrics that align with RLER. The Evidence Grounding Score (EGS) measures frame legality and coverage: for a candidate $o_i$ with cited frames $K_i$, we define $v_i$ as the fraction of valid, in-range, non-duplicate indices, and $u_i$ as the Jaccard overlap with the consensus set $\bar{K}$ from the election step, then $\mathrm{EGS}(o_i){=}\tfrac{1}{2}(v_i{+}u_i)$ averaged over candidates. The Transparency Index (TI) instantiates the bounded length preference $\tau(o_i){=}\sin^{2}(\pi\,\widetilde{L}(o_i))$ and reports the mean value across correctly parsed traces. The Redundancy Ratio (RR) captures surface repetition using fixed $n$-grams: $\rho(o_i){=}1{-}\tfrac{|\mathcal{U}_n(o_i)|}{T_n(o_i)}$, and we report the mean $\rho$ (lower is better). 

\noindent\textbf{Implementation Details.}
We adopt Qwen2.5-VL-7B-Instruct~\cite{Qwen2.5-VL} and train on NVIDIA A100 40GB GPUs. Only language-model parameters are updated; the vision encoder and projection are frozen. We use LoRA ($r{=}8$, $\alpha{=}16$, dropout $0.05$) and AdamW (lr $1{\times}10^{-5}$, weight decay $0.1$, grad clip $1.0$). GRPO uses group size $G{=}4$, clipping $\epsilon_{\text{clip}}{=}0.2$, and KL coefficient $\beta{=}0.04$ for two epochs on Video-R1-260k~\cite{feng2025videor1}. Training samples 16 frames per video uniformly. At inference we enable RLER-Inference with budget cap $K_{\max}{=}8$ and a fixed early-stop rule shared across benchmarks. Candidates are diversified by a temperature grid $\{0.2,0.7,0.9\}$ with top-$p{=}0.9$ and distinct seeds, plus center and four-corner crops at $0.9$ scale with mild blur ($\sigma{=}0.5$) and $\pm10\%$ brightness; one conservative run is always included. Standard tasks use 32 frames; long videos are subsampled at 1 fps. If the schema is missing, we fall back to answer-level aggregation; all other settings follow official protocols. Training Curves are in Appendix. 

Unless noted, reward weights are $w_{\text{acc}}{=}0.1$, $w_{\text{fmt}}{=}0.1$, $w_{\text{fs}}{=}0.2$, $w_{\text{tt}}{=}0.3$, $w_{\text{ar}}{=}0.3$ in $R_{\text{total}}=\sum_k w_k R_k$. The frame-sensitive score $s_{\mathrm{fs}}(o)$ follows Eq.~\eqref{eq:kqs-compact} on parsed \texttt{<keyframes>} with deduplication and range checks. Think-transparency maps the reasoning length $L(o)$, clipped to $[64,1024]$ tokens, to $\widetilde{L}(o)\in[0,1]$ and uses $\tau(o)=\sin^{2}(\pi\widetilde{L}(o))$. Anti-repetition uses 5-grams over \texttt{<think>}: $\rho(o)=1-\frac{|\mathcal{U}_3(o)|}{T_3(o)}$ and $R_{\text{ar}}=-\rho(o)$. The format reward is $R_{\text{fmt}}(o)=\mathcal{F}(o)$ with mandatory \texttt{<think>}, \texttt{<answer>}, \texttt{<keyframes>}; accuracy uses exact match after canonicalization. GRPO advantages use z-score with $\epsilon=10^{-8}$; the reference policy is the initial snapshot and $\pi_{\theta_{\text{old}}}$ is refreshed once per step. Early stopping triggers when the evidence-weighted margin exceeds $\delta{=}0.08$ and the mean confidence of the leading cluster exceeds $\gamma{=}0.4$; confidence $c_i$ is the average token probability over the last $25\%$ of $o_i$. Election is evidence-weighted with at most one maximal and one minimal $S_i$ trimmed per answer cluster.

\begin{figure*}[t]
    \centering
    \includegraphics[width=\linewidth]{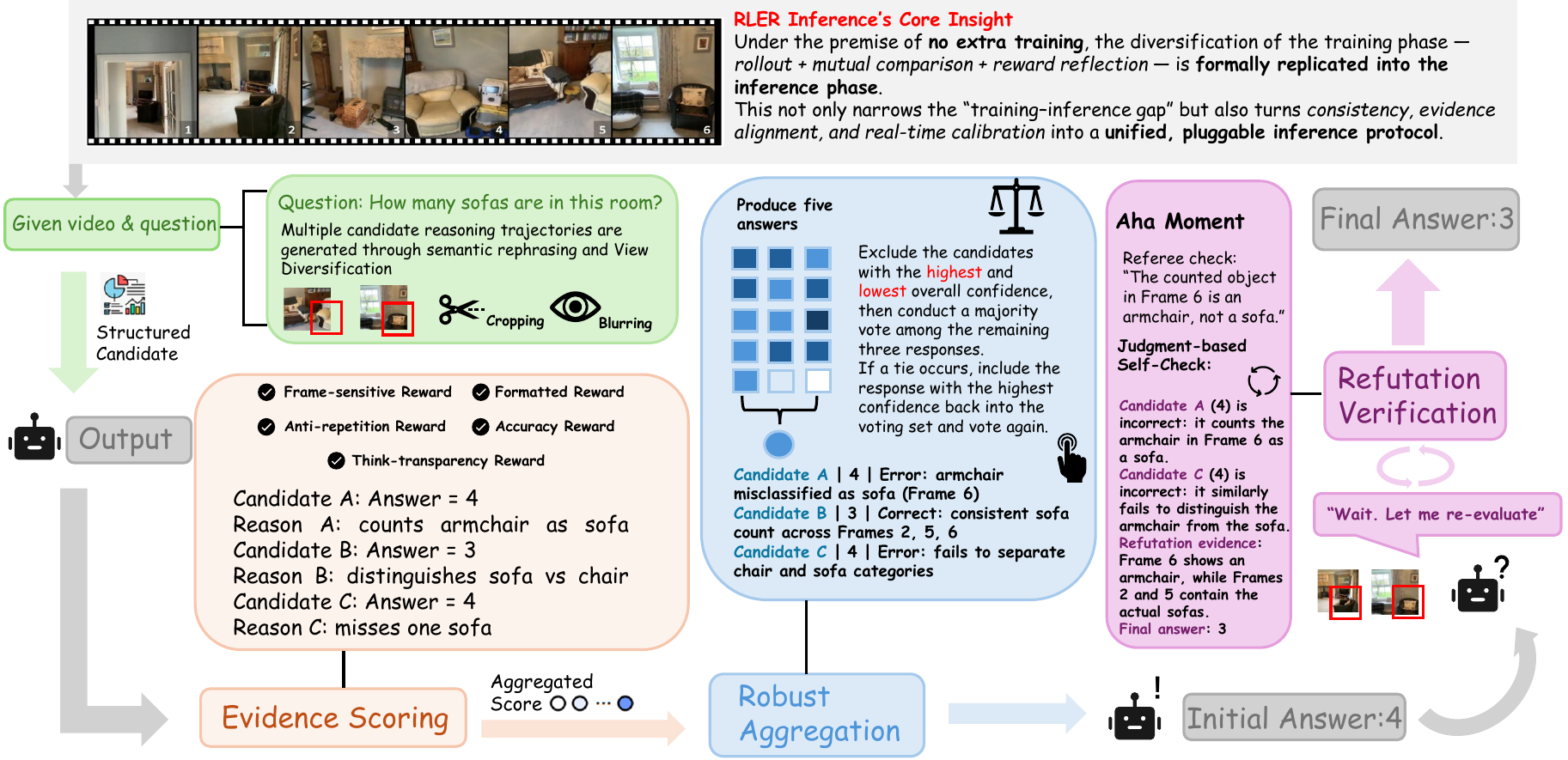}
    \vspace{-7mm}
    \caption{A case study show how RLER uses diverse inputs to form structured candidates, scores evidence, aggregates robustly, and performs refutation verification to revise the initial answer and deliver the final result.}
    \label{fig:case}
    \vspace{-5mm}
\end{figure*}

\begin{figure}[t]
    \centering
    \includegraphics[width=\linewidth]{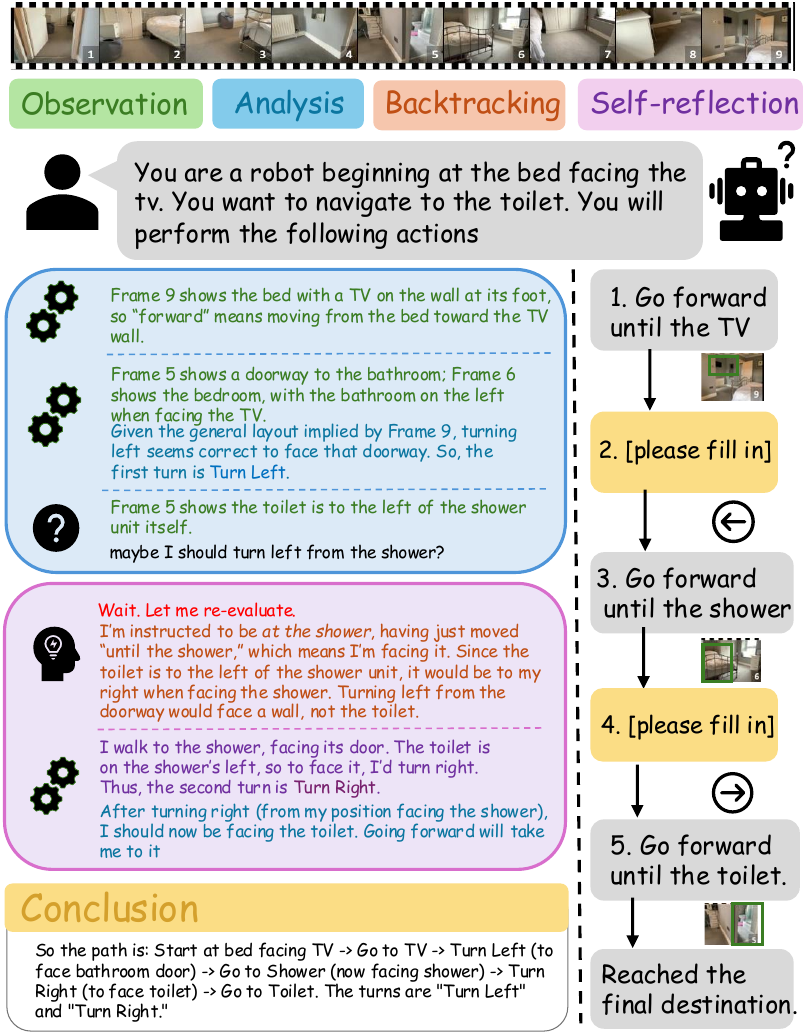}
    \vspace{-6.5mm}
    \caption{An example of emergence elicited by RLER-Training. The {\color{red} red} text marks the ``aha moment'', where the model identifies an internal conflict and initiates a re-evaluation.}
    \vspace{-6mm}
    \label{fig:aha}
\end{figure}

\subsection{Main Result}
RLER delivers consistent gains across all three benchmark families in Table~\ref{tab:video_results_grouped}. On video reasoning, it sets the best open-source results with 43.3\% on VSIBench and 54.2\% on VideoMMMU, surpassing recent RL-based models and improving markedly over the base model. On general video understanding, it achieves 68.5\% on VideoMME without subtitles, 76.2\% on TempCompass, 72.9\% on MVBench, and 57.5\% on WildVideo; these are the highest open-source numbers reported, and on VideoMME it even exceeds the proprietary GPT-4o entry. On long video understanding, it reaches 50.7\% on LVBench and 63.0\% on LongVideoBench, outperforming all reported open-source and RL-based baselines. The pattern is consistent: once training shapes structured evidence and inference elects by evidence, accuracy rises across diverse settings that stress cross-frame reasoning, mixed perception–reasoning skills, and long temporal horizons.

\noindent\textbf{``Aha moment" Emergence in RLER-Training.}
During RLER-Training process, we observe a distinct emergence phenomenon we term the ``aha moment." We define this as a discrete event where the model explicitly recognizes a flaw in its ongoing hypothesis and initiates a corrective thought process. As shown in Figure~\ref{fig:aha}, this is captured by the model's interjection, ``Wait. Let me re-evaluate.'' Crucially, these self-initiated reversals furnish actionable cues for RLER-Inference: the same pattern is detected by the referee-style self-check to challenge a provisional winner and, when warranted, trigger reweighting or one-shot resampling, thereby linking training-time emergence to inference-time verification.

\noindent\textbf{Case Study.}
As shown in Figure~\ref{fig:case}, a single-pass model outputs an incorrect count due to conflating visually similar furniture. RLER generates diversified candidates, parses each into answer, cited keyframes, and reasoning, and scores them by evidence consistency, transparency, and non-redundancy. Robust aggregation then down-weights outliers and elects the most evidence-aligned hypothesis. A brief referee self-check triggers an “aha moment,” where the candidate revisits its cited evidence and corrects the miscount, yielding the final correct answer 3. This example highlights RLER’s strengths: structured evidence from training, principled arbitration at inference, and targeted self-revision that turns ambiguous scenes into reliable, interpretable decisions.

\subsection{Ablation Study}

\noindent\textbf{Training Ablations.} As shown in Table~\ref{tab:ablate_train}, under matched data and optimization, RLER is best on both datasets (43.3\% on VSIBench; 54.2\% on VideoMMMU). Removing the frame-sensitive reward causes the largest drop on VSIBench to 41.0\%, while removing think-transparency or anti-repetition yields smaller but steady declines. Without RLER-Inference reaches 41.7\% on VSIBench and 52.5\% on VideoMMMU, indicating that reward shaping alone strengthens reasoning and evidence quality, though some ambiguous cases remain without election. Replacing GRPO with SFT records 39.2\% on VSIBench and 49.8\% on VideoMMMU. Despite having the required format, it lacks the reward signals that activate stronger reasoning, which shows that RLER-Training is better suited to complex reasoning and that the full dual paradigm is most effective.

\begin{table}[h]
\centering
\small
\caption{\textbf{Training ablations.} We ablate each training reward of RLER-Training while keeping the inference pipeline fixed, and also evaluate two asymmetric settings: w/o RLER-Inference means without RLER-Inference setting, w/o GRPO (w/ SFT) means replacing GRPO training with SFT.}
\vspace{-2mm}
\begin{adjustbox}{width=\linewidth,center}
\setlength{\tabcolsep}{3.6pt}
\begin{tabular}{lcc}
\toprule
Setting & VSIBench $\uparrow$ & VideoMMMU $\uparrow$ \\
\midrule
Qwen2.5-VL-7B (Baseline) & 37.4 & 47.4 \\
\rowcolor{blue!10} RLER (Ours)  & \textbf{43.3} & \textbf{54.2} \\
\midrule
\;\;w/o frame-sensitive & 41.0 & 52.1 \\
\;\;w/o think-transparency & 41.9 & 52.7 \\
\;\;w/o anti-repetition & 42.1 & 53.0 \\
\;\;w/o RLER-Inference & 41.7 & 52.5 \\
\;\;w/o GRPO (w/ SFT) & 39.2 & 49.8 \\
\bottomrule
\end{tabular}
\end{adjustbox}
\label{tab:ablate_train}
\vspace{-2mm}
\end{table}

\begin{table}[h]
\centering
\caption{\textbf{Inference ablations.} We remove one RLER-Inference component at a time while fixing other setting. “w/o diversity input” disables all input diversification, so $K{=}1$. “w/o evidence weight” replaces evidence-weighted election with equal-weighted majority voting. “w/o outlier trimming” keeps extreme-score candidates. “w/o tran. \& red.” means only frame scoring. “w/o referee self-check” skips the final critique. “w/o RLER-Training” applies the pipeline to a model trained without RLER-Training; missing structured fields trigger answer-level aggregation. Avg $K$ is the mean number of evaluated candidates under $K_{\max}{=}8$.}
\vspace{-2mm}
\begin{adjustbox}{width=\linewidth,center}
\setlength{\tabcolsep}{3.6pt}
\begin{tabular}{lccc}
\toprule
Setting & MVBench $\uparrow$ & LVBench $\uparrow$ & Avg $K$ \\
\midrule
Qwen2.5-VL-7B & 67.5 & 42.0 & -- \\
\rowcolor{blue!10} RLER (Ours) & \textbf{72.9} & \textbf{50.7} & 3.1 \\
\midrule
\;\;w/o diversity input & 69.1 & 46.5 & 1.0 \\
\;\;w/o evidence weight & 70.6 & 48.3 & 3.1 \\
\;\;w/o outlier trimming & 71.4 & 49.2 & 3.2 \\
\;\;w/o tran. \& red. & 70.2 & 47.9 & 3.1 \\
\;\;w/o referee self-check & 71.8 & 49.1 & 3.0 \\
\;\;w/o RLER-Training  & 68.1 & 42.5 & 3.0 \\
\bottomrule
\end{tabular}
\end{adjustbox}
\label{tab:ablate_infer}
\vspace{-6mm}
\end{table}

\noindent\textbf{Inference Ablations.}
As shown in Table~\ref{tab:ablate_infer}, RLER reaches 72.9\% on MVBench and 50.7\% on LVBench with Avg $K{=}3.1$. Removing diversity input causes the largest drop, to 69.1\% and 46.5\% at $K{=}1.0$, showing the value of alternative views. Replacing evidence weighting with majority vote reduces accuracy to 70.6\% and 48.3\% at a similar budget, indicating that grounded evidence should carry more influence than raw counts. Disabling outlier trimming yields 71.4\% and 49.2\% with Avg $K{=}3.2$, and using frames only gives 70.2\% and 47.9\%, both reflecting weaker consensus. Omitting the referee self-check produces 71.8\% and 49.1\%, a small but consistent loss. Applying the pipeline without RLER-Training gives 68.1\% and 42.5\%, since reward-shaped outputs provide stronger and more consistent evidence for election.

\begin{table}[t]
\centering
\caption{\textbf{Compute–quality tradeoff.} We compare single sample ($K{=}1$), RLER-I(nference) with early stopping ($K_{\max}{=}8$), and the fixed budget ($K{=}8$).}
\vspace{-2mm}
\begin{adjustbox}{width=\linewidth,center}
\setlength{\tabcolsep}{4pt}
\begin{tabular}{lcccc}
\toprule
Setting & VSIBench & VideoMMMU & MVBench & Avg $K$ \\
\midrule
Single sample  & 41.0 & 52.0 & 69.1 & 1.0 \\
\rowcolor{blue!10} RLER (Ours) & \textbf{43.3} & \textbf{54.2} & \textbf{72.9} & 3.1 \\
Fixed budget & 43.9 & 54.6 & 73.3 & 8.0 \\
\bottomrule
\end{tabular}
\end{adjustbox}
\label{tab:budget_quality}
\vspace{-3mm}
\end{table}

\noindent\textbf{Compute–Quality Tradeoff.}
As shown in Table~\ref{tab:budget_quality}, With the trained model and $K_{\max}{=}8$ fixed, dynamic RLER-I raises VSIBench from 41.0\% to 43.3\%, VideoMMMU from 52.0\% to 54.2\%, and MVBench from 69.1\% to 72.9\% using on average $K{=}3.1$. Increasing to a fixed $K{=}8$ adds only 0.4–0.6\% at about $2.6\times$ the sampling cost. This shows that evidence aligned election reaches consensus quickly on most examples; extra samples contribute little and sometimes add redundant candidates. MVBench exhibits the largest step from single sample to dynamic, while the move from dynamic to $K{=}8$ has clear diminishing returns.

\begin{table}[t]
\centering
\small
\caption{\textbf{Evidence–performance mirroring.} EGS measures keyframe validity and coverage; TI measures transparency from a bounded-length preference; RR is $n$-gram redundancy. }
\vspace{-2mm}
\begin{adjustbox}{width=0.9\linewidth,center}
\setlength{\tabcolsep}{2.4pt}
\begin{tabular}{lcccc}
\toprule
Setting & Acc $\uparrow$ & EGS $\uparrow$ & TI $\uparrow$ & RR $\downarrow$ \\
\midrule
Qwen2.5-VL-7B & 37.4 & 0.41 & 0.31 & 0.27 \\
\rowcolor{blue!10} RLER (Ours) & \textbf{43.3} & \textbf{0.72} & \textbf{0.64} & \textbf{0.11} \\
\midrule
\;\;w/o frame-sensitive & 41.0 & 0.58 & 0.62 & 0.12 \\
\;\;w/o think-transparency & 41.9 & 0.70 & 0.48 & 0.12 \\
\;\;w/o anti-repetition & 42.1 & 0.71 & 0.63 & 0.18 \\
\bottomrule
\end{tabular}
\end{adjustbox}
\label{tab:mirror_evidence}
\vspace{-5mm}
\end{table}

\noindent\textbf{Evidence–Performance Mirroring.}
As shown in Table~\ref{tab:mirror_evidence}, Full RLER attains 43.3\% with EGS 0.72, TI 0.64, and RR 0.11. Dropping the frame-sensitive reward reduces EGS to 0.58 and accuracy to 41.0\%, reflecting the need for cross-frame grounding. Dropping the think-transparency reward lowers TI to 0.48 with accuracy 41.9\%, showing the value of readable structure. Dropping the anti-repetition reward raises RR to 0.18 with accuracy 42.1\%, indicating that redundancy dilutes useful evidence. The metric that shifts most in each row tracks the accuracy change, confirming that training shapes verifiable evidence and inference prefers candidates that present it coherently.

\vspace{-1mm}
\section{Conclusion}
\vspace{-1mm}
We addressed the unreliability of single-pass video reasoning by introducing RLER, a dual paradigm that learns to produce evidence and then reasons by electing among it. RLER-Training uses 3 novel rewards to shape structured and verifiable traces that expose key evidence, raise information density, and improve transparency. RLER-Inference consumes these traces, extracts and scores evidence, and performs an robust evidence-weighted election. Across video reasoning, general understanding, and long-horizon benchmarks, RLER improves accuracy and evidence quality in a compute-efficient manner, showing that explicit evidence in learning and evidence-aligned arbitration at inference are complementary and necessary.

\noindent \textbf{Acknowledgement.} This work is supported by the Young Scientists Fund of the Hunan Natural Science Foundation (Grant No.2024JJ6474), the Youth Independent Innovation Science Fund Project of NUDT (Grant No.ZK24-08).

{
    \small
    \bibliographystyle{ieeenat_fullname}
    \bibliography{main}
}

\end{document}